\documentclass[letterpaper, 10 pt, conference]{ieeeconf}  

\IEEEoverridecommandlockouts                              

\usepackage{comment}
\usepackage{tabularx}
\usepackage{booktabs}
\usepackage{subcaption}
\usepackage{caption} 
\usepackage{mathtools}
\usepackage{amsmath}    
\usepackage{amssymb}    
\usepackage{graphicx}   
\usepackage{bm}         
\usepackage{url}
\usepackage{graphicx}      

\usepackage{xcolor}
\usepackage{float}      
\usepackage{multirow}
\usepackage{bbm}
\usepackage{booktabs} 
\usepackage{extdash}
\usepackage{multirow}
\usepackage{array}     
\usepackage{boldline}  
\usepackage{siunitx}

\usepackage[sorting=none,style=ieee]{biblatex}
\usepackage{caption}
\title{\LARGE \bf
Application-Specific Component-Aware Structured Pruning of \\Deep Neural Networks in Control via Soft Coefficient Optimization
}

\author{Ganesh Sundaram$^{*}$$^{1}$, Jonas Ulmen$^{*}$$^{1}$, Amjad Haider$^{*}$$^{1}$ and Daniel Görges$^{*}$$^{1}$
\thanks{*Authors have contributed equally}
\thanks{$^{1}$Department of Electrical and Computer Engineering, RPTU University Kaiserslautern-Landau, Germany. E-mail:
        {\tt\small \{ganesh.sundaram, jonas.ulmen, ahaider, daniel.goerges\}@rptu.de}}
}

\begin{document}

\maketitle
\thispagestyle{empty}
\pagestyle{empty}

\begin{abstract}

Deep neural networks (DNNs) offer significant flexibility and robust performance. This makes them ideal for building not only system models but also advanced neural network controllers (NNCs). However, their high complexity and computational needs often limit their use. Various model compression strategies have been developed over the past few decades to address these issues. These strategies are effective for general DNNs but do not directly apply to NNCs. NNCs need both size reduction and the retention of key application-specific performance features. In structured pruning, which removes groups of related elements, standard importance metrics often fail to protect these critical characteristics. In this paper, we introduce a novel framework for calculating importance metrics in pruning groups. This framework not only shrinks the model size but also considers various application-specific constraints. To find the best pruning coefficient for each group, we evaluate two approaches. The first approach involves simple exploration through grid search. The second utilizes gradient descent optimization, aiming to balance compression and task performance. We test our method in two use cases: one on an MNIST autoencoder and the other on a Temporal Difference Model Predictive Control (TDMPC) agent. Results show that the method effectively maintains application-relevant performance while achieving a significant reduction in model size.
\end{abstract}

\section{Introduction}
The deployment of DNNs has significantly advanced machine learning capabilities, enabling improved representation learning, generalization, and task performance across diverse applications. As the demand for richer and more informative input observations increases, neural networks have grown in both architectural complexity and input dimensionality. Modern neural network models incorporate high-dimensional modalities, including text, images, and audio, to extract comprehensive environmental information. This development has widened the gap between large, powerful models created in research environments and the practical constraints of real-world deployments, such as limited computational resources, memory, and energy efficiency~\cite{liu2025survey}.

Model compression techniques, particularly pruning, have rapidly evolved to address these challenges. Pruning reduces the size and computational burden of neural networks by removing redundant or less important parameters. Despite their effectiveness, a central challenge persists: compressing models aggressively without sacrificing performance or accuracy, especially when neural network behavior is highly application-specific.

An illustrative example is found in networks designed to learn compressed representations of high-dimensional data, such as images. These networks distill essential features into compact latent spaces, enabling efficient downstream learning and inference. Pruning latent space-based architectures, however, presents significant challenges. As latent space compression increases, the sensitivity of learned representations to pruning also rises, making it challenging to maintain task-relevant behaviors without precise control over the pruning process. Similarly, NNCs trained to converge can achieve optimal performance on complex control tasks. Subsequent pruning to reduce computational requirements, however, often leads to significant performance degradation, resulting in unexpected or even catastrophic outcomes. This degradation may occur whether pruning is applied in the actual state space or in learned latent representations, with the latter introducing a more complex, cascading problem in which errors in the latent space compound with mistakes in the control policy.

Pruning methods are generally categorized as structured or unstructured pruning~\cite{li2023model}. Structured pruning eliminates entire architectural components, such as filters, channels, or layers, resulting in genuine reductions in model size and memory usage and delivering performance gains on standard hardware without specialized accelerators. In contrast, unstructured pruning removes individual weights, often producing sparse matrices that are less effective without specialized hardware. Structured pruning libraries typically define groups of structurally coherent elements and remove these groups based on importance metrics. Most frameworks use absolute value or Euclidean norm-based criteria to evaluate group importance and prune those with the lowest magnitude. However, these approaches risk discarding parameters or groups that, despite having low magnitude, are essential for maintaining performance, particularly in application-specific or highly compressed contexts. This paper addresses these challenges by exploring performance-preserving structured pruning strategies that provide finer control over the pruning process. The focus is on maintaining task-relevant behaviors, particularly in highly compressed latent-space models where conventional pruning criteria may be inadequate. Novel methods for evaluating group importance and optimizing pruning decisions are proposed to ensure that model compression does not compromise critical performance metrics.

The structure of this paper is as follows. Section~\ref{sec:related_works} reviews prior work on neural network compression and NNC architectures. Section~\ref{sec:experiment} details the proposed methodology, including component-aware pruning, soft coefficient assignment, and optimization strategies, demonstrated on MNIST autoencoder and TD-MPC use cases. Section~\ref{sec:results} presents the experimental results and discusses key findings. Section~\ref{sec:conclusion} concludes the paper and outlines directions for future research.

\vspace{-0.1cm}
\section{Related Work}
\label{sec:related_works}

Research on compression techniques for NNC applications is still in its early stages of development. Reducing the complexity of DNNs in this field is a critical area of investigation, motivated by the significant computational and memory requirements that impede deployment on resource-constrained devices~\cite{cheng2024survey}. Among various compression strategies, structured network pruning is extensively studied. A primary challenge in structured pruning is establishing effective criteria for identifying the least essential components. The most prevalent heuristic is magnitude-based pruning, which removes structures with the smallest $\ell$-norms. Frameworks such as Structurally Prune Anything (SPA)~\cite{wangStructurallyPruneAnything2024} and the methods by Hussien et al.~\cite{hussien2024small} utilize this straightforward yet practical approach. However, a key limitation is that the magnitude of weight does not always reflect a structure's true functional significance~\cite{castells2024ld}. More advanced methods incorporate additional information to address this issue. For example, Molchanov et al.~\cite{molchanov2019importance} applied Taylor expansions to estimate the impact of removing a neuron, and Zhuang et al.~\cite{zhuang2020neuron} introduced a polarization regularizer that leverages Batch Normalization scaling factors as proxies for neuron importance. Other approaches employ Fisher information to assess significance~\cite{scholl2021information}. Although these data-driven methods are more principled, they can be computationally intensive and are often tailored to specific architectures, which limits their generalizability. Another category of methods formulates pruning as an explicit optimization problem. The CHITA framework~\cite{benbaki2023fast} models pruning as a combinatorial optimization task, utilizing Iterative Hard Thresholding to optimize the pruned structure and the remaining weights jointly. Additional approaches include framing neuron selection as a submodular optimization problem~\cite{el2022data} or employing probabilistic methods with learnable masks for each filter~\cite{li2021filter}. These methods often introduce complex optimization landscapes or rely on theoretical assumptions that may not hold in practice, and scalability remains a significant concern.

Reinforcement Learning (RL) is often formulated as an optimal control problem, leading to RL methods that closely resemble NNCs~\cite{jonker_model-based_nodate}. Latent-space approaches, including those in the Dreamer family~\cite{hafner2024masteringdiversedomainsworld}, V-JEPA 2~\cite{assran2025vjepa2selfsupervisedvideo}, and the TD-MPC family~\cite{hansen2022temporaldifferencelearningmodel}, have demonstrated state-of-the-art performance across a wide range of RL tasks. In contrast, control techniques that operate directly in the state space rather than in a latent space provide formal stability guarantees. Notable examples are neural Lyapunov control methods~\cite{zhou2022neurallyapunovcontrolunknown} and neural control barrier functions~\cite{so2024train}, which employ Lyapunov stability theory and barrier certificates to ensure provable safety and stability in nonlinear control systems.

A primary limitation of current pruning techniques is their simplistic allocation of the sparsity budget, often applying a uniform ratio to all layers or relying on basic heuristics. This approach fails to achieve an optimal balance between model performance and size. In this paper, we present a more principled framework to address this issue. First, a functional grouping method tailored for multi-component architectures partitions the network into meaningful structural blocks. Second, a novel gradient-based optimizer determines the optimal pruning coefficient for each group. Through numerical gradient estimation, the optimizer efficiently explores the continuous space of coefficients to identify configurations that maximize performance for a specified sparsity target. This approach yields a criterion-agnostic and automated method for allocating sparsity across a network's structure. As an extension, control-specific metrics could be employed to distribute the sparsity budget according to each layer's relevance to control performance. For example, layers responsible for generating control commands or processing critical sensor inputs should be pruned conservatively, whereas layers focused on feature extraction or redundant transformations can accommodate higher sparsity. This control-aware allocation strategy has the potential to maintain closed-loop performance while achieving compression ratios that are comparable to or better than those of generic methods.

\section{Experimental Setup}
\label{sec:experiment}

To establish a concrete context for the proposed methodology, this section introduces the experimental setup and use cases. This structure illustrates key concepts and motivates the technical framework presented in Section~\ref{sec:methodology} through practical examples. The methodology is subsequently developed with explicit reference to these use cases, demonstrating how application-specific requirements inform the design of the pruning framework.

\subsection{Component-Specific Structured Pruning}

In complex network structures, dependencies often emerge among groups of parameters, which necessitate the simultaneous pruning of these groups. Frameworks such as \texttt{Torch-Pruning} enable systematic pruning by constructing dependency graphs that explicitly represent the interdependencies among structurally coupled parameters as pruning groups. This approach supports informed pruning decisions that maintain network functionality while maximizing compression benefits~\cite{Fang_2023_CVPR}. In previous work~\cite{sundaram2025enhancedpruningstrategymulticomponent}, the dependency graph framework was extended to address Multi-Component Neural Architectures (MCNAs) by generating individual dependency graphs for each component instead of a single monolithic graph. This extension facilitates the identification of both component-specific pruning groups and inter-component coupling groups. The resulting decomposed representation enhances flexibility in pruning decisions, allowing for the targeted selection of pruning groups and the effective isolation of sensitive architectural or functional elements that must be preserved.

\subsection{Latent Representation Models}

Complex systems cannot generally be fully understood using only simple sensory inputs. Modern approaches now utilize high-dimensional data, such as images and videos, to construct more comprehensive internal models of system behavior. NNCs reflect this shift, moving from post-hoc parameter reduction toward principled representation learning. The core idea is to use an encoder that maps high-dimensional observations into a compact, low-dimensional latent space. This latent space captures only the essential dynamics relevant to control tasks. By reducing data at this early stage, before any control policy is applied, the encoder eliminates redundant information from the outset.

However, compressing models that rely on latent encodings introduces distinct challenges. The latent representation is itself a compressed, information-rich bottleneck that is carefully learned during training. Aggressive compression of the encoder can disrupt this learned mapping, causing severe information loss. Moreover, any distortion in the latent space propagates to all downstream components that depend on it, leading to unpredictable effects on system performance. Therefore, the application-specific framework must account for these constraints and preserve the integrity of the latent space, maintaining its properties throughout the compression process.

\subsection{Use Cases}

The proposed methodology is evaluated through two complementary use cases. The first involves a toy problem utilizing an autoencoder on the MNIST dataset. Autoencoders represent the most basic architecture employing latent space encoding, which is fundamental to advanced NNCs. Their visual reconstruction outputs enable an intuitive assessment of pruning effects, facilitating a direct comparison between the proposed framework and conventional approaches. The second use case implements TD-MPC, a state-of-the-art model-based RL framework that combines latent representations with model predictive control, demonstrating the method's applicability to complex control tasks that require maintaining stability and convergence during compression.

\subsubsection{MNIST Autoencoder}

\begin{figure}[h]
\captionsetup{width=\linewidth}
    \centering
    \includegraphics[width=\linewidth]{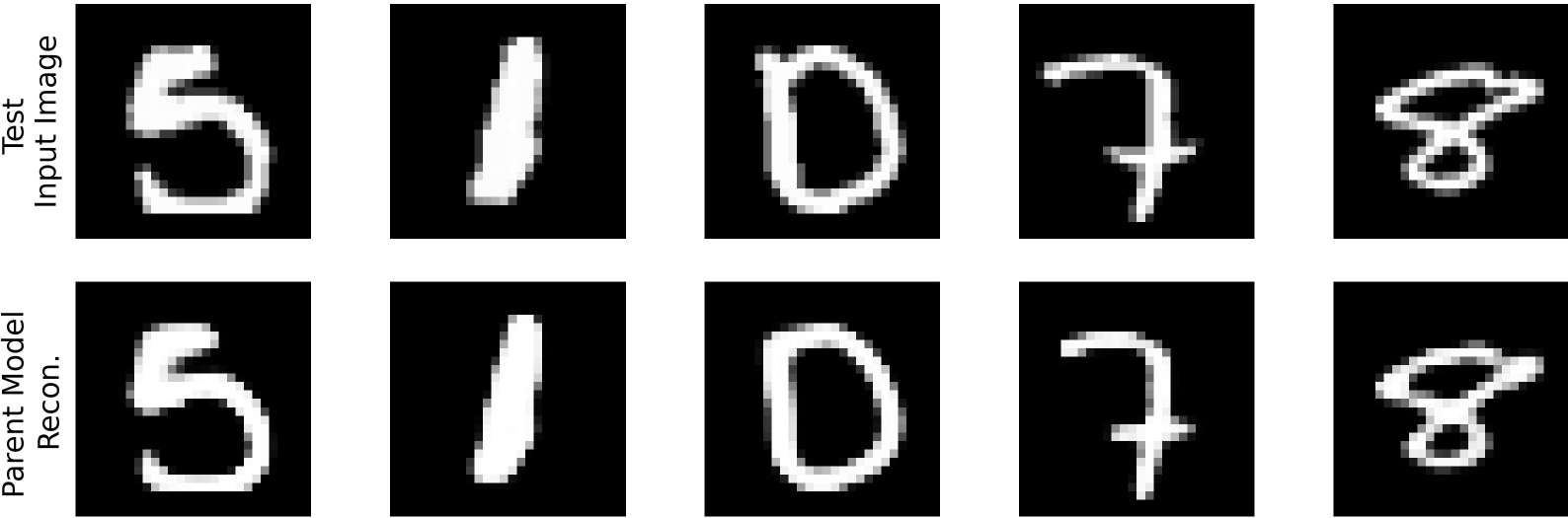}
    \caption{Reconstruction quality of the baseline (unpruned) model. The top row displays the original input images from the test set, and the bottom row shows their corresponding reconstructions.}
    \label{fig:parent_model_reconstructions}
\end{figure}

The MNIST dataset is a standard benchmark containing grayscale images of handwritten digits, widely used to evaluate image models and representation learning capabilities~\cite{zhu2018classification}. An autoencoder compresses these inputs into a low-dimensional latent representation using an \emph{encoder} and then reconstructs the original image from this compressed code using a \emph{decoder}. This architecture learns compact features that preserve essential visual content. The model is trained end-to-end to minimize reconstruction loss. Table~\ref{tab:autoencodermodel_params} summarizes the key characteristics of the baseline autoencoder architecture. Figure~\ref{fig:parent_model_reconstructions} shows representative reconstruction results from this unpruned baseline model.

\begin{table}[h]
\centering
\caption{Baseline autoencoder model characteristics}
\begin{tabular}{@{}lc@{\hspace{1.5em}}lc@{}}
\toprule
\textbf{Parameter}  & \textbf{Value}            & \textbf{Parameter}        & \textbf{Value} \\
\midrule
Input Size          & $28 \times 28$ pixels     & Latent Dim.               & 256 \\
Components          & Encoder and Decoder       & Parameters                & 2,312,992 \\
Layers/Component    & 2                         & Model Size                & \SI{8.84}{MB} \\
Peak PSNR           & $22.34\,\mathrm{dB}$      & Final MSE                 & 0.002238 \\
\bottomrule
\bottomrule
\end{tabular}
\label{tab:autoencodermodel_params}
\end{table}

\begin{table}[!h]
\captionsetup{width=\linewidth}
\caption{Pruning group decomposition for MNIST Autoencoder showing component-specific and coupling groups with their constituent modules and parameter counts}
\label{tab:pruning_groups_ae}
\centering
\scriptsize
\sisetup{group-separator={,}}
    \begin{tabular}{@{}llccc@{}}
        \toprule
        \textbf{Group Type} & \textbf{Component} & \textbf{Group} & \textbf{Modules} & \textbf{Parameters} \\ \midrule
        \multirow{4}{*}{\textit{Component-specific}} & \multirow{2}{*}{Encoder} & 1 & 8 & \num{202148} \\
         & & 2 & 4 & \num{339560} \\ \cmidrule(l){2-5}
         & \multirow{2}{*}{Decoder} & 1 & 8 & \num{339692} \\
         & & 2 & 4 & \num{202280} \\ \midrule
        \textit{Coupling} & Encoder-Decoder & 1 & 1 & \num{99584} \\
        \bottomrule
        \bottomrule
    \end{tabular}
\end{table}

Table~\ref{tab:pruning_groups_ae} presents a detailed breakdown of the groups created using the component-aware dependency graph method for the MNIST Autoencoder model, including their constituent modules and parameter counts.

\subsubsection{TD-MPC – Balancing Inverted Pendulum}

TD-MPC is a model-based reinforcement learning framework that implements model predictive control (MPC) within a learned latent state space~\cite{hansen2022temporaldifferencelearningmodel}. This approach simultaneously trains a latent dynamics model and a terminal value function using temporal-difference learning. Subsequently, it plans optimized action sequences over short horizons in the compact latent representation. The TD-MPC agent consists of five neural network components: an \emph{encoder} $h_{\theta}$ that maps pixel observations $s_t$ to latent states $z_t$; a \emph{dynamics} model $f_{\theta}$ that predicts the next latent state $z_{t+1}$ from the current state $z_t$ and action $a_t$; a \emph{reward} model $r_{\theta}$ that estimates immediate rewards based on state-action pairs; a terminal \emph{value function} $V_{\phi}$ that predicts long-term returns from a latent state; and a \emph{policy} $\pi_{\psi}(a \mid z)$ trained to imitate the MPC planner for rapid inference.

The proposed method is evaluated on the classic inverted pendulum swing-up task from OpenAI Gym~\cite{brockman2016openai}. In this task, the agent learns from raw grayscale pixel observations to swing the pendulum from its downward resting position to an upright state and maintain balance. The encoder compresses high-dimensional images into a compact latent representation. At the same time, the learned world model predicts future transitions within this space, facilitating efficient MPC-based planning for both swing-up and stabilization. Table~\ref{tab:model_params} summarizes the model features and training details. Table~\ref{tab:pruning_groups_tdmpc} presents a detailed breakdown of the groups formed, including the constituent modules and parameter counts for the trained TD-MPC model.

\begin{table}[h]
\centering
\captionsetup{width=\linewidth}
\caption{TD-MPC model characteristics and training specifications}
\begin{tabular}{@{}lc@{\hspace{1em}}lc@{}}
\toprule
\textbf{Parameter}  & \textbf{Value}            & \textbf{Parameter} & \textbf{Value} \\
\midrule
Input Size          & $28 \times 28$ pixels     & Latent Dim.       & 50 \\
Components          & Encoder,                  & Episode Reward    & 858.6 \\
                    & Dynamics, Reward,         & Episode Length    & 125 steps \\
                    & Value, Policy             & Total Parameters  & 1{,}548{,}360 \\
Layers/Comp.        & 3                         & Total FLOPs       & $4.12 \times 10^{8}$ \\
Hidden Units        & 512                       & Model Size        & \SI{12.4}{MB} \\
\bottomrule
\bottomrule
\end{tabular}
\label{tab:model_params}
\end{table}

\vspace{-0.2cm}

\begin{table}[h]
\captionsetup{width=\linewidth}
\caption{Pruning group decomposition for TD-MPC showing component-specific and coupling groups with their constituent modules and parameter counts}
\label{tab:pruning_groups_tdmpc}
\centering
\scriptsize
\sisetup{group-separator={,}}
    \begin{tabular}{@{}llccc@{}}
        \toprule
        \textbf{Group Type} & \textbf{Component} & \textbf{Group} & \textbf{Modules} & \textbf{Parameters} \\ \midrule
        \multirow{12}{*}{\textit{Comp.-specific}} & 
            \multirow{3}{*}{Encoder} & 1 & 1 & \num{25632} \\ & & 2 & 1 & \num{9248} \\ & & 3 & 1 & \num{9248} \\ \cmidrule(l){2-5} & 
            \multirow{1}{*}{Dynamics} & 1 & 1 & \num{262656} \\ \cmidrule(l){2-5} & 
            \multirow{1}{*}{Reward} & 1 & 1 & \num{262656} \\ \cmidrule(l){2-5} &
            \multirow{1}{*}{Pi} & 1 & 1 & \num{262656} \\ \cmidrule(l){2-5} &
            \multirow{1}{*}{Q1} & 1 & 1 & \num{262656} \\ \cmidrule(l){2-5} &
            \multirow{1}{*}{Q2} & 1 & 1 & \num{262656} \\   \midrule
        \multirow{2}{*}{\textit{Coupling}} & 
             \multirow{1}{*}{Encoder-Pi} & 1 & 2 & \num{26112} \\ \cmidrule(l){2-5} &
             \multirow{1}{*}{EncoderPi-Dyn.Rew.Q1Q2} & 1 & 5 & \num{369152} \\ 
        \bottomrule
        \bottomrule
    \end{tabular}
\end{table}

\vspace{-0.2cm}
\section{Application-Specific Target Sparsity Pruning} 
\label{sec:methodology}

\subsection{Limitations of Norm-Based Importance Calculation}

Simple pruning techniques, such as the binary decision method (i.e., retaining or eliminating an entire group) and conventional norm-based importance metrics, often lead to significant performance degradation. To address this limitation, we introduce more flexible and tunable pruning coefficients to each identified group, enabling fine-grained control over the pruning intensity.

While component-aware pruning enhances grouping strategies, the approach for determining the extent of pruning for each group largely follows traditional practices. Typically, groups are ranked based on importance metrics calculated from the $\ell_1$ or $\ell_2$ norm of their constituent weights. Groups with the lowest norm values receive the lowest importance scores and are subsequently pruned. This approach is generally adequate for applications where the primary objective is to achieve specific compression ratios by reducing model size.

However, conventional importance metrics do not address application-specific constraints beyond basic model compression. For instance, specific prediction tasks require a minimum level of accuracy, and specialized domains, such as control systems, necessitate the preservation of critical properties, including stability, convergence, and robustness, following model compression. Consequently, norm-based importance measures cannot guarantee that pruned models maintain optimal or near-optimal performance under domain-specific requirements. The group with the lowest norm score may still contain essential information required to meet these criteria. This limitation highlights a fundamental weakness in the current approach to determining importance scores.

\subsection{Assignment of Soft Coefficients to Pruning Groups}

The proposed application-aware framework enables partial parameter removal by assigning a tunable soft coefficient, $c_i \in [0, 1]$, to each pruning group. This coefficient determines the proportion of parameters to be pruned from each group, where $c_i = 0$ indicates no pruning and $c_i = 1$ indicates complete removal. To prevent structural disconnection that may result from pruning a group entirely, the upper pruning limit is set to 0.95. This 95\% maximum ensures the retention of some neurons, thereby maintaining the information flow throughout the network. The framework provides fine-grained control over pruning intensity for each group based on its impact on overall model performance. The soft coefficient method supports the identification of optimal configurations within the pruning-performance trade-off space and mitigates the abrupt performance degradation often observed with conventional importance metrics. As indicated in Tables~\ref{tab:pruning_groups_ae} and~\ref{tab:pruning_groups_tdmpc}, the MNIST autoencoder utilizes five coefficients, whereas the TDMPC model utilizes ten. Table~\ref{tab:coefficient_notation} summarizes the notation for the various soft coefficients used in both use cases.

\begin{table}[h]
\centering
\caption{Soft coefficients assigned to the pruning groups}
\label{tab:coefficient_notation}
\begin{tabular}{@{}llc@{}r@{$\;\in\;$}l@{}}
\toprule
\textbf{Model} & \textbf{Type} & \textbf{Notation} & \multicolumn{2}{c}{\textbf{Range}} \\
\midrule
\multirow{2}{*}{Autoencoder} & Component-specific & $c^{\text{ae}}_{i_\text{comp}}$ & $i_\text{comp}$ & $\{1,\ldots,4\}$ \\
 & Coupling & $c^{\text{ae}}_{i_\text{coup}}$ & $i_\text{coup}$ & $\{1\}$ \\
\midrule
\multirow{2}{*}{TD-MPC} & Component-specific & $c^{\text{tdmpc}}_{j_\text{comp}}$ & $j_\text{comp}$ & $\{1,\ldots,8\}$ \\
 & Coupling & $c^{\text{tdmpc}}_{j_\text{coup}}$ & $j_\text{coup}$ & $\{1,2\}$ \\
\bottomrule
\bottomrule
\end{tabular}
\end{table}

\vspace{-0.5cm}

\subsection{Performance Evaluation Function}

Pruned models remain viable only if they satisfy specific application requirements. To ensure compliance, the pruning process must incorporate mechanisms for monitoring and enforcing these constraints. This requirement is addressed by defining a performance metric, which is evaluated after each pruning operation—a step referred to as \textit{performance evaluation}. The metric may consist of any application-specific measure that quantifies task-relevant behaviors, such as reconstruction loss, system stability, prediction accuracy, or other domain-specific indicators. In addition to assessment, the metric serves a dual role: evaluating the impact of each pruning configuration and guiding the search for optimal pruning coefficients.

For the MNIST autoencoder, the Peak Signal-to-Noise Ratio (PSNR) is employed as a standard metric to quantify the impact of pruning on image reconstruction quality. Higher PSNR values correspond to lower distortion between the original and reconstructed images, indicating improved performance. For TD-MPC, total episode reward serves as the metric, where a higher reward demonstrates that the agent maintains the pendulum upright and within a specified angle range.

\subsection{Pruning Coefficients Search}

The subsequent step involves identifying the optimal set of pruning coefficients, $c_i$ and $c_j$, which determine the extent of pruning for each group in both use case examples. This objective is addressed using two distinct approaches.

\paragraph{\textbf{Grid Search}}The grid search approach systematically evaluates all possible coefficient combinations within a predefined search space. For each group coefficient, a linearly spaced grid is constructed from 0.0 to 0.95 with $n$ intermediate points. For $m$ parameter groups, this process generates an $m \times n$ grid of coefficient combinations.

The grid search operates through a two-stage filtering process. First, each coefficient combination is evaluated to identify those that achieve the target pruning magnitude. Combinations resulting in significantly different overall model sparsity are filtered out, with a practical tolerance of $\pm x\%$ applied to retain viable options. Second, the remaining coefficient sets are used to prune the model, and each resulting configuration is assessed using the predefined performance evaluator. Excluding unsuitable coefficient sets in advance reduces unnecessary computations and processing time. The combination yielding the highest performance metric is ultimately selected as optimal, ensuring that the final configuration achieves both the required sparsity and maximum model performance.

\paragraph{\textbf{Constrained Optimization}}Although grid search is a straightforward optimization method, it exhibits significant limitations. It is subject to the curse of dimensionality, rendering it computationally impractical as the number of groups \(m\) or the grid resolution increases. Furthermore, its discrete nature restricts the search to predefined points, potentially missing superior solutions between these points. To overcome these challenges, the pruning task is reformulated as a continuous constrained optimization problem. This reformulation enables the use of advanced search algorithms that can efficiently identify the optimal set of pruning coefficients. The problem is defined as
\begin{subequations}
\label{eq:pruning_optimization_general}
    \begin{alignat}{2}
        &\!\min_{\mathbf{c}} \quad & & J(\mathbf{c}) \label{eq:pruning_optimization_general:objective} \\
        &\:\:\mathrm{s.t.} \quad 
        & & \rho - \varepsilon \leq \frac{\sum_{i=1}^{m} c_i s_i}{\|\theta\|_0} \leq \rho + \varepsilon \label{eq:pruning_optimization_general:constraint1} \\
        & & & \mathbf{c} \in [0, 0.95]^m \label{eq:pruning_optimization_general:constraint2}
    \end{alignat}
\end{subequations}

In this formulation, \(J(\mathbf{c})\) denotes the application-specific objective function that quantifies the performance of the model pruned using the coefficient vector \(\mathbf{c} = [c_1, \dots, c_m]^T\). The primary constraint enforces the target sparsity of the model. The term \(c_i s_i\) represents the number of parameters pruned from group \(i\), where \(s_i\) is the size of the group.

The expression $\sum c_i s_i / \|\theta\|_0$ quantifies the achieved total sparsity of the model, where $\|\theta\|_0 = \sum_{i=1}^{m} s_i$ denotes the total number of parameters in the baseline model. The achieved sparsity is constrained to remain within a tolerance window $[\rho - \varepsilon, \rho + \varepsilon]$ around the desired sparsity target $\rho$, with $\varepsilon$ specifying the tolerance. This formulation enables the optimizer to determine the optimal distribution of pruning intensities across all groups, thereby maximizing performance while strictly maintaining the overall sparsity budget.

A primary technical challenge arises from the inherent non-differentiability of the pruning process, which renders standard gradient-based optimizers ineffective, as they operate in continuous space. This non-differentiability arises from the conversion of pruning coefficients from a continuous domain into discrete binary masks, which determine the proportion of weights in each group to be pruned by setting them to zero. The masking step introduces abrupt, step-like changes. Consequently, slight variations in coefficients can result in sudden transitions from retained to pruned weights, thereby disrupting the smooth gradient flow required for analytical differentiation. To overcome this limitation, a custom optimization framework is employed.

The proposed methodology utilizes a composite objective function that integrates the performance metric with a quadratic penalty for deviations from the target sparsity. As this objective is not analytically differentiable, its gradient is estimated numerically using the central finite difference method. This approach explores the loss landscape to determine the direction of steepest descent. Subsequently, the coefficients are updated using standard gradient descent, augmented with a momentum term to guide the optimization process.

In this custom implementation, the optimizer considers the coefficients for all pruning groups as a single parameter vector. Unlike conventional gradient descent methods that may employ random restart strategies to avoid local minima, this approach utilizes a direct and deterministic strategy. The pruning process is treated as monotonic, with all coefficients initialized at zero and incrementally increased until the target sparsity is reached. The numerical gradient calculation independently assesses the impact of each coefficient on the total loss, which includes both model performance and the sparsity constraint. The update step then applies the complete gradient vector to adjust all coefficients simultaneously, each according to its respective gradient component.

\section{Results and Discussion}
\label{sec:results}

This section presents the experimental results of the proposed application-aware pruning framework applied to both use cases. To establish performance baselines, we compare our results against those obtained using standard structured pruning from the \texttt{PyTorch} library \texttt{torch.nn.utils.prune}.

All experiments were conducted on a PC equipped with an Intel Core i\num{7}-{8650}U CPU and \num{16}{GB} RAM. The target sparsity ($\rho$) was fixed at 20\% with a tolerance of $\varepsilon = \pm 1\%$ for both coefficient search methods, allowing for a direct comparison between approaches. This uniform sparsity target was maintained across both the MNIST autoencoder and TD-MPC use cases to ensure consistent evaluation conditions.

\subsection{Use Case I: MNIST Autoencoder}

\begin{figure}[h]
\captionsetup{width=\linewidth}
    \centering
    \includegraphics[width=\linewidth]{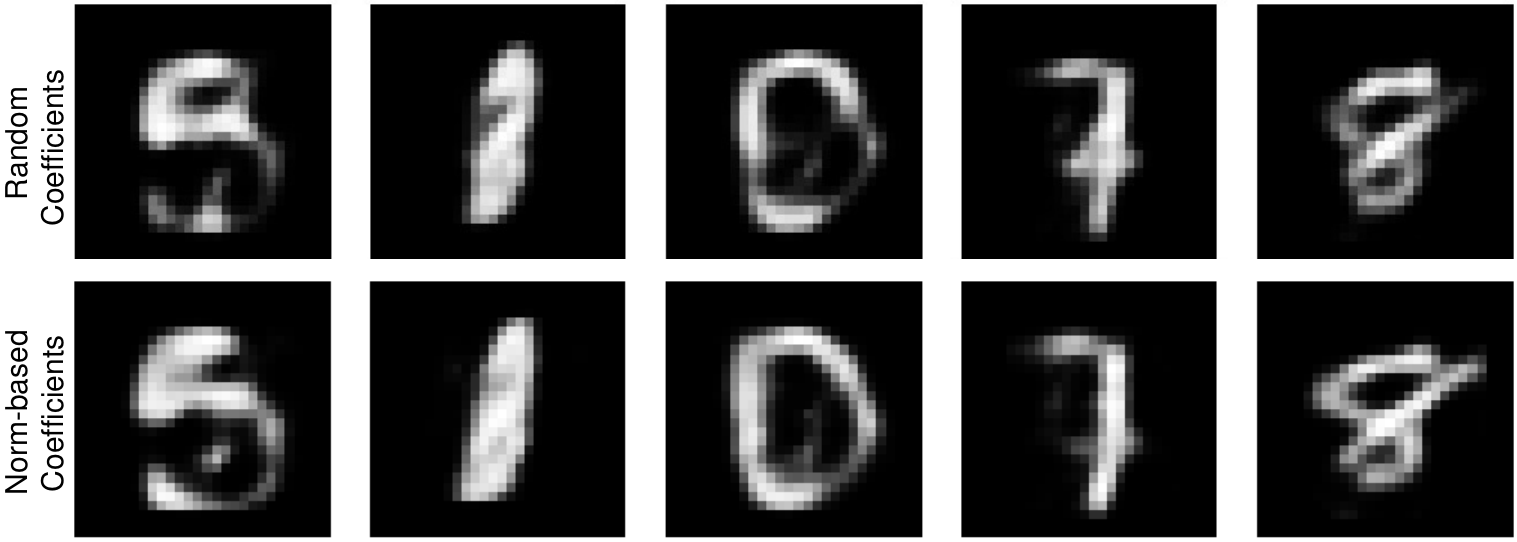}
    \caption{Comparison of reconstruction quality for a model pruned to 20\% sparsity using random (top) and norm-based (bottom) coefficient selection with \texttt{PyTorch}’s standard pruning library.}
    \label{fig:pruned_model_random_vs_normbased}
\end{figure}

To establish baseline performance, two importance criteria were evaluated: random pruning, which arbitrarily removes parameters, and magnitude-based pruning, which eliminates parameters with the smallest $\ell_1$ or $\ell_2$ norm values. Figure~\ref{fig:pruned_model_random_vs_normbased} compares the reconstruction quality at 20\% sparsity for both methods, demonstrating the significant impact of the importance metric selection on visual performance.

The visual results reveal a substantial performance difference. Randomly assigned pruning coefficients result in severe image degradation, rendering some digits unrecognizable. In contrast, the norm-based approach, which prioritizes pruning groups with smaller weight norms, produces reconstructions with higher fidelity and identifiable digits. Although superior to the random baseline, the norm-based heuristic does not guarantee an optimal solution.

For the exhaustive \emph{grid search}, the search space comprised five pruning groups, each with ten intermediate points between 0 and 0.95, resulting in $10^5$ distinct combinations. The search proceeded in two stages:

\paragraph{Sparsity Filtering} The initial set of combinations was filtered to retain only those that satisfied the target model's sparsity. This step yielded 252 viable candidate solutions. 
\paragraph{Performance Evaluation} Each candidate set was used to prune the model, and the resulting performance was evaluated using the PSNR metric.

\begin{figure}[h]
\captionsetup{width=\linewidth}
  \begin{subfigure}[t]{\linewidth}
    \centering
    \includegraphics[width=\linewidth]{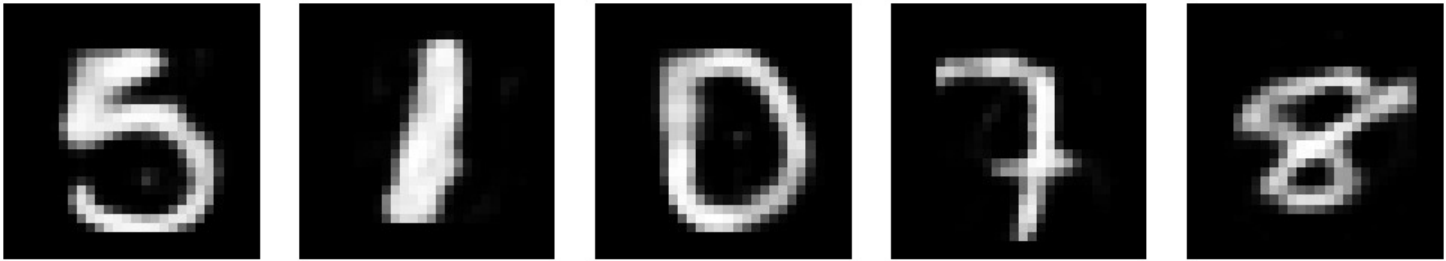}
    \caption{Grid search}
  \end{subfigure}
  \vspace{-0.1cm}
  \begin{subfigure}[t]{\linewidth}
    \centering
    \includegraphics[width=\linewidth]{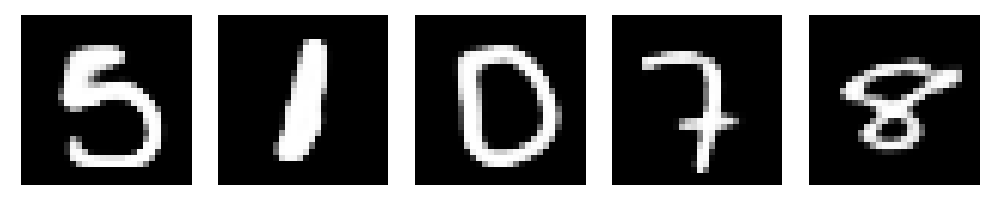}
    \caption{Gradient descent optimization}
  \end{subfigure}
    \caption{Reconstruction quality comparison for models pruned to 20\% sparsity using (a) grid search and (b) gradient descent optimization.}
\label{fig:reconstruction_pruned_model}
\end{figure}

\begin{table}[h]
    \captionsetup{width=\linewidth}
    \caption{Performance comparison of grid search and gradient descent optimization methods}
    \label{tab:comparison_gs_gd}
    \centering
    
    \small 
    \sisetup{round-mode=places, round-precision=4}

    \begin{tabular}{@{} l S[table-format=3.4] S[table-format=3.4] @{}}
    \toprule
    
    \textbf{Metric} & {\textbf{Grid Search}} & {\textbf{Gradient Descent}} \\
    \midrule
    PSNR ($\mathrm{dB}$)            & 22.13                 & \bfseries 25.34 \\
    Target Sparsity (\%)            & 20.00                 & 20.00 \\
    Achieved Sparsity (\%)          & 19.07                 & \bfseries 20.80 \\
    Pruned Model MSE                & 0.008633              & \bfseries 0.008528 \\
    Baseline Model MSE              & 0.002238              & 0.002238 \\
    $\Delta$MSE                     & 0.006395              & \bfseries 0.006290 \\
    \textbf{Runtime ($\mathrm{s}$)} & 408.85                & \bfseries 16.73 \\
    \midrule
    
    \multicolumn{3}{@{}l}{\textbf{Optimal Coefficient Set} ($\mathbf{c}$)} \\
    \cmidrule(r){1-3}
    Grid Search             & \multicolumn{2}{l}{[0.527, 0.633, 0.633, 0.527, 0.316]} \\
    Gradient Descent        & \multicolumn{2}{l}{[0.569, 0.568, 0.655, 0.523, 0.495]} \\
    \bottomrule
    \bottomrule
    \end{tabular}
\end{table}

Table~\ref{tab:comparison_gs_gd} presents the performance metrics of pruned models with optimal configurations. Both methods achieve the target sparsity while maintaining reconstruction quality. However, \emph{gradient descent optimization} outperforms \emph{grid search} by identifying a more effective set of coefficients. Specifically, it achieves a higher PSNR ($\SI{25.34}{dB}$ compared to $\SI{22.13}{dB}$), as demonstrated by the sharper and more accurate reconstruction in Fig.~\ref{fig:reconstruction_pruned_model}. Additionally, it attains sparsity levels closer to the target and reduces runtime by 96\% (\SI{16.73}{s} compared to \SI{408.85}{s}). Gradient descent also achieves a higher peak PSNR than the baseline model's $\SI{22.34}{dB}$, as shown in Table~\ref{tab:autoencodermodel_params}, despite a reduction in parameters. This improvement indicates that application-aware pruning can enhance model performance by serving as an effective regularizer, removing redundant or noisy parameters while preserving structures essential to the task. Regularization is particularly advantageous in latent-space architectures, where parameter redundancy may introduce noise into the compressed representation. These results underscore the effectiveness of gradient-based optimization for application-aware pruning of neural networks.

\subsection{Use Case II: TD-MPC – Balancing Inverted Pendulum}

\begin{figure}[h]
\captionsetup{width=\linewidth}
    \centering
    \includegraphics[width=\linewidth]{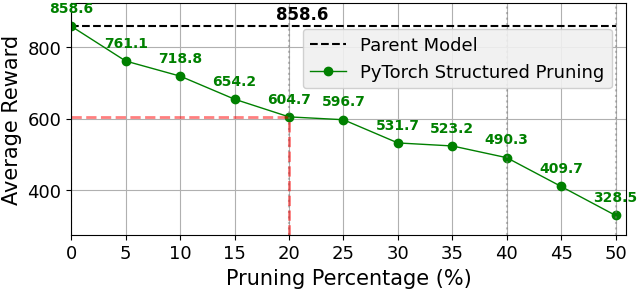}
    \caption{Performance degradation of TD-MPC under \texttt{PyTorch} structured pruning using magnitude-based importance. The red dashed line indicates 20\% pruning, at which point baseline performance declines to 604.7, demonstrating severe degradation.}
    \label{fig:parent_model_rewards}
\end{figure}

To evaluate the effectiveness of baseline methods, structured magnitude-based pruning was applied to the TD-MPC architecture, and performance was measured as sparsity increased. Figure~\ref{fig:parent_model_rewards} illustrates a pronounced decline in average episode reward as the pruning percentage rises. The reward at 20\% pruning, which serves as the comparison level for the proposed method, is indicated by a red dashed line. These findings suggest that, in contrast to other contexts, the baseline approach results in a significant loss of control performance (over 30\% reduction at 20\% pruning), underscoring the limitations of conventional pruning criteria in complex, multi-component control architectures that utilize latent representations.

The TD-MPC architecture, which consists of ten pruning groups, presents substantial computational challenges for an exhaustive \emph{grid search}. Evaluating ten intermediate points between 0 and 0.95 for each coefficient yields $10^{10}$ possible combinations, necessitating extensive computational time. While reducing the number of intermediate points $n$ would lower computational requirements, it would also decrease the granularity of achievable pruning ratios and reduce the probability of identifying optimal coefficient configurations. This scalability constraint underscores the necessity of employing gradient descent optimization in complex, multi-component architectures.

During \emph{gradient descent optimization}, each iteration applies the current set of pruning coefficients to the parent model and evaluates the pruned model over a complete episode to obtain the total episode reward. This reward signal directly represents the agent's control performance, providing immediate feedback on the effects of pruning decisions on task execution. Table~\ref{tab:grad_desc_tdmpc} summarizes the performance metrics and the optimal configuration identified through this optimization process.

\begin{table}[h]
    \sisetup{group-separator={,}}
    \captionsetup{width=\linewidth}
    \caption{Performance of the gradient descent optimization for TD-MPC}
    \label{tab:grad_desc_tdmpc}
    \centering   
    \small 

    \begin{tabular}{@{} l c @{}} 
    \toprule
    
    \textbf{Metric}                     & {\textbf{Value}}      \\
    \midrule
    Episode Reward                      & 720.08                \\
    Target Sparsity (\%)                & 20.00                 \\
    Achieved Sparsity (\%)              & 19.70                 \\
    Number of Iterations                & 64                    \\
    Model size (params)                 & \num{1258740}         \\
    \textbf{Runtime ($\mathrm{s}$)}     & 46.22                 \\
    \midrule
    
    \multicolumn{2}{@{}l}{\textbf{Optimal Coefficient Set} ($\mathbf{c}$)} \\
    \cmidrule(r){1-2}
    \textit{Component-specific:}    & \multicolumn{1}{c}{[0.0531, 0.0586, 0.0600, 0.2040,}\\
                                    & \multicolumn{1}{c}{0.2286, 0.2370, 0.2441, 0.2258]}\\

    \textit{Coupling:} & \multicolumn{1}{c}{[0.0, 0.0]} \\

    \bottomrule
    \bottomrule
    \end{tabular}
\end{table}

The results indicate that application-aware pruning outperforms conventional methods. At 20\% sparsity, the model from \emph{gradient descent optimization} achieves an episode reward of 720.08, compared to the baseline's 604.7, representing a 19.1\% improvement in absolute performance. This substantial gain supports the conclusion that application-specific coefficient optimization preserves task-critical structures that conventional magnitude-based metrics do not capture.

Analysis of the optimal coefficient set provides critical insights into the optimizer's behavior. The optimizer completely avoids pruning the coupling groups $c^{\text{tdmpc}}_{j_\text{coup}}$, thereby preserving essential inter-component connections that maintain information flow between the encoder, policy, and world model components. The optimizer also demonstrates restraint when pruning the encoder's component-specific groups $c^{\text{tdmpc}}_{\text{comp},j}$ for $j \in \{1,2,3\}$, which are responsible for transforming high-dimensional pixel observations into latent encodings necessary for accurate state estimation in model predictive control. In contrast, the optimizer applies substantially higher pruning to downstream components $c^{\text{tdmpc}}_{\text{comp},j}$ for $j \in \{4,\ldots,8\}$, where redundancy is more acceptable. This emergent behavior, identified automatically through optimization rather than manual design, demonstrates the framework's capacity to respect architectural dependencies and allocate pruning pressure according to functional importance. The \SI{46.22}{s} runtime further highlights the practical efficiency of gradient-based optimization for complex multi-component NNCs.

\section{Conclusion and Future Work}
\label{sec:conclusion}

Conventional structured pruning methods that rely on heuristic-based strategies, particularly those using group weight norms, often fail to identify critical pruning groups. This limitation results in notable performance degradation in NNCs and in applications with sparsity-sensitive components, such as latent spaces. The common assumption that smaller groups are inherently less critical is not universally valid, since even minimal groups can encode essential information for task performance. Consequently, heuristic approaches often yield suboptimal pruning decisions, which can reduce model accuracy and potentially destabilize an NNC, ultimately compromising its practical utility.

In the paper, a principled framework has been proposed to address these limitations by assigning soft, learnable coefficients to each pruning group and applying optimization techniques to determine their optimal values. Two complementary methods have been evaluated: grid search and gradient descent, both of which use application-specific performance evaluation functions tailored to task requirements. Experimental results have demonstrated that both optimization techniques significantly outperform baseline approaches employing standard structured pruning with random and norm-based selection, thereby validating the effectiveness of the optimization-driven framework.

While grid search is straightforward to implement, its computational demands increase rapidly with higher grid dimensions, rendering it impractical for larger networks. For the MNIST autoencoder, both methods achieved comparable reconstruction quality. However, gradient descent yielded a 24-fold reduction in runtime, demonstrating superior computational efficiency. In the more complex TDMPC scenario, grid search was entirely infeasible, further emphasizing its scalability limitations. These results establish gradient-based optimization as the practical choice for real-world applications where computational resources and time constraints are critical.

Future work will extend this framework in two complementary directions. First, the importance criteria will be enhanced by incorporating multiple theoretical constraints, including stability guarantees and sensitivity analysis, to strengthen the framework's applicability to safety-critical NNCs. Second, importance identification will be integrated directly into the training process using Bayesian inference, Fisher information matrices, or gradient accumulation techniques. This integration will enable the immediate determination of pruning group importance upon training completion, eliminating the need for a separate post-training analysis and streamlining the deployment pipeline for task-specific compressed models.

\printbibliography

@InProceedings{Fang_2023_CVPR,
    author    = {Fang, Gongfan and Ma, Xinyin and Song, Mingli and Mi, Michael Bi and Wang, Xinchao},
    title     = {DepGraph: Towards Any Structural Pruning},
    booktitle = {Proceedings of the 2023 IEEE/CVF Conference on Computer Vision and Pattern Recognition},
    month     = {06},
    year      = {2023},
    pages     = {16091-16101}
}

@inproceedings{sundaram2025enhancedpruningstrategymulticomponent,
    title        = {Enhanced Pruning Strategy for Multi-Component Neural Architectures Using Component-Aware Graph Analysis},
    author       = {Ganesh Sundaram and Jonas Ulmen and Daniel G\"orges},
    booktitle    = {Proceedings of the IFAC Joint Conference on Computers, Cognition and Communication (J3C)},
    month        = {September},
    year         = {2025}
}

@article{liu2025survey,
  title={A survey of model compression techniques: Past, present, and future},
  author={Liu, Defu and Zhu, Yixiao and Liu, Zhe and Liu, Yi and Han, Changlin and Tian, Jinkai and Li, Ruihao and Yi, Wei},
  journal={Proceedings of the Frontiers in Robotics and AI},
  volume={12},
  pages={1518965},
  year={2025},
  publisher={Frontiers Media SA}
}

@article{li2023model,
  title={Model compression for deep neural networks: A survey},
  author={Li, Zhuo and Li, Hengyi and Meng, Lin},
  journal={Proceedings of the Computers},
  volume={12},
  number={3},
  pages={60},
  year={2023},
  publisher={MDPI}
}

@online{wangStructurallyPruneAnything2024,
  title={Structurally Prune Anything: Any Architecture, Any Framework, Any Time},
  author={Xun Wang and John Rachwan and Stephan G{\"u}nnemann and Bertrand Charpentier},
  year = {2023},
  doi = {10.48550/arXiv.2403.18955},
}

@article{brockman2016openai,
  title={{OpenAI} gym},
  author={Brockman, Greg and Cheung, Vicki and Pettersson, Ludwig and Schneider, Jonas and Schulman, John and Tang, Jie and Zaremba, Wojciech},
  year={2016},
  journal={arXiv:1606.01540},
  archivePrefix={arXiv},
  primaryClass={cs.LG}
}

@inproceedings{hansen2022temporaldifferencelearningmodel,
  title={Temporal difference learning for model predictive control},
  author={Nicklas Hansen and Xiaolong Wang and Hao Su},
  booktitle={Proceedings of the 39th International Conference on Machine Learning},
  pages={8387--8406},
  year={2022}
}

@article{zhu2018classification,
  title={Classification of MNIST handwritten digit database using neural network},
  author={Zhu, Wan},
  journal={Proceedings of the Research School of Computer Science. Australian National University, Acton, ACT},
  volume={2601},
  year={2018}
}

@article{cheng2024survey,
  title={A survey on deep neural network pruning: Taxonomy, comparison, analysis, and recommendations},
  author={Cheng, Hongrong and Zhang, Miao and Shi, Javen Qinfeng},
  journal={Proceedings of the IEEE Transactions on Pattern Analysis and Machine Intelligence},
  year={2024}
}

@article{hussien2024small,
  title={Small Contributions, Small Networks: Efficient Neural Network Pruning Based on Relative Importance},
  author={Hussien, Mostafa and Afifi, Mahmoud and Nguyen, Kim Khoa and Cheriet, Mohamed},
  journal={arXiv preprint arXiv:2410.16151},
  year={2024}
}

@article{zhuang2020neuron,
  title={Neuron-level structured pruning using polarization regularizer},
  author={Zhuang, Tao and Zhang, Zhixuan and Huang, Yuheng and Zeng, Xiaoyi and Shuang, Kai and Li, Xiang},
  journal={Advances in Neural Information Processing Systems},
  volume={33},
  pages={9865--9877},
  year={2020}
}

@inproceedings{molchanov2019importance,
  title={Importance estimation for neural network pruning},
  author={Molchanov, Pavlo and Mallya, Arun and Tyree, Stephen and Frosio, Iuri and Kautz, Jan},
  booktitle={Proceedings of the 2019 IEEE/CVF Conference on Computer Cision and Cattern Recognition},
  pages={11264--11272},
  year={2019}
}

@article{scholl2021information,
  title={The information theory of developmental pruning: Optimizing global network architectures using local synaptic rules},
  author={Scholl, Carolin and Rule, Michael E and Hennig, Matthias H},
  journal={PLoS Computational Biology},
  volume={17},
  number={10},
  pages={e1009458},
  year={2021},
  publisher={Public Library of Science San Francisco, CA USA}
}

@article{el2022data,
  title={Data-efficient structured pruning via submodular optimization},
  author={El Halabi, Marwa and Srinivas, Suraj and Lacoste-Julien, Simon},
  journal={Proceedings of the Advances in Neural Information Processing Systems},
  volume={35},
  pages={36613--36626},
  year={2022}
}

@inproceedings{benbaki2023fast,
  title={Fast as chita: Neural network pruning with combinatorial optimization},
  author={Benbaki, Riade and Chen, Wenyu and Meng, Xiang and Hazimeh, Hussein and Ponomareva, Natalia and Zhao, Zhe and Mazumder, Rahul},
  booktitle={Proceedings of the 40th International Conference on Machine Learning},
  pages={2031--2049},
  year={2023}
}

@inproceedings{li2021filter,
  title={Filter pruning via probabilistic model-based optimization for accelerating deep convolutional neural networks},
  author={Li, Qinghua and Li, Cuiping and Chen, Hong},
  booktitle={Proceedings of the 14th ACM International Conference on Web Search and Data Mining},
  pages={653--661},
  year={2021}
}

@inproceedings{castells2024ld,
  title={Ld-pruner: Efficient pruning of latent diffusion models using task-agnostic insights},
  author={Castells, Thibault and Song, Hyoung-Kyu and Kim, Bo-Kyeong and Choi, Shinkook},
  booktitle={Proceedings of the 2024 IEEE/CVF Conference on Computer Vision and Pattern Recognition},
  pages={821--830},
  year={2024}
}

@misc{hafner2024masteringdiversedomainsworld,
      title={Mastering Diverse Domains through World Models}, 
      author={Danijar Hafner and Jurgis Pasukonis and Jimmy Ba and Timothy Lillicrap},
      year={2024},
      eprint={2301.04104},
      archivePrefix={arXiv},
      primaryClass={cs.AI},
}

@misc{assran2025vjepa2selfsupervisedvideo,
      title={V-JEPA 2: Self-Supervised Video Models Enable Understanding, Prediction and Planning}, 
      author={Mido Assran and Adrien Bardes and David Fan and Quentin Garrido and Russell Howes and Mojtaba and Komeili and Matthew Muckley and Ammar Rizvi and Claire Roberts and Koustuv Sinha and Artem Zholus and Sergio Arnaud and Abha Gejji and Ada Martin and Francois Robert Hogan and Daniel Dugas and Piotr Bojanowski and Vasil Khalidov and Patrick Labatut and Francisco Massa and Marc Szafraniec and Kapil Krishnakumar and Yong Li and Xiaodong Ma and Sarath Chandar and Franziska Meier and Yann LeCun and Michael Rabbat and Nicolas Ballas},
      year={2025},
      eprint={2506.09985},
      archivePrefix={arXiv},
      primaryClass={cs.AI}, 
}

@article{jonker_model-based_nodate,
  title={Model-based reinforcement learning: A survey},
  author={Moerland, Thomas M and Broekens, Joost and Plaat, Aske and Jonker, Catholijn M and others},
  journal={Proceedings of the Foundations and Trends in Machine Learning},
  volume={16},
  number={1},
  pages={1--118},
  year={2023},
  publisher={Now Publishers, Inc.}
}

@misc{zhou2022neurallyapunovcontrolunknown,
      title={Neural Lyapunov Control of Unknown Nonlinear Systems with Stability Guarantees}, 
      author={Ruikun Zhou and Thanin Quartz and Hans De Sterck and Jun Liu},
      year={2022},
      eprint={2206.01913},
      archivePrefix={arXiv},
      primaryClass={eess.SY}, 
}

@inproceedings{so2024train,
  title={How to train your neural control barrier function: Learning safety filters for complex input-constrained systems},
  author={So, Oswin and Serlin, Zachary and Mann, Makai and Gonzales, Jake and Rutledge, Kwesi and Roy, Nicholas and Fan, Chuchu},
  booktitle={Proceedings of the 2024 IEEE International Conference on Robotics and Automation},
  pages={11532--11539},
  year={2024}
}

\end{document}